\documentclass{bmvc2k}
\usepackage{booktabs}
\usepackage{amsfonts}

\title{ExSinGAN: Learning an Explainable Generative Model From a Single Image}

\addauthor{Zicheng Zhang}{zhangzicheng19@mails.ucas.ac.cn}{1}
\addauthor{Congying Han$^*$}{hancy@ucas.ac.cn}{1}
\addauthor{Tiande Guo}{tdguo@ucas.ac.cn}{1}
\addinstitution{
	University of Chinese Academy of Science, Beijing, China
}

\runninghead{Zhang, Han, Guo}{ExSinGAN For SINGLE IMAGE GENERATION}

\def\eg{\emph{e.g}\bmvaOneDot}

\def\etal{\emph{et al}\bmvaOneDot}

\begin{document}
	
	\maketitle

	\begin{abstract}
		Generating images from a single sample has attracted extensive attention recently. In this paper, we formulate the problem as sampling from the conditional distribution of a single image, and propose a hierarchical framework that simplifies the learning of the intractable conditional distribution through the successive learning of the distributions about structure, semantics, and texture, making the generative model more comprehensible compared with previous works. On this basis, we design ExSinGAN composed of three modular GANs for learning an explainable generative model from a given image, where the modular GANs model the distributions about structure, semantics, and texture successively. ExSinGAN is learned not only from the internal patches of the given image as the previous works did, but also from the external prior obtained by the GAN inversion technique. Benefiting from the appropriate combination of internal and external information, ExSinGAN has a more powerful capability of generation and competitive generalization ability for image manipulation tasks. 
	\end{abstract}

	\section{Introduction}\label{Introduction}
	\par Recently, single image generative models get more and more attention. The groundbreaking work~\cite{Shaham2019SinGANLA} by  Shaham \etal   proposed a pyramid of fully convolutional GANs~\cite{goodfellow2014generative} named SinGAN to learn the internal distribution of patches within the single image. SinGAN is an unconditional generative model that acquires noise inputs and produces diverse syntheses but only needs the one-shot training image. Due to the impressive performance and extensible architecture, SinGAN has been the paradigm of single image generation, which is improved~\cite{Gur2020HierarchicalPV, Hinz2020ImprovedTF, Xu2020PositionalEA, Mahendren2021DiverseSI, Granot2021DropTG} and applied to other fields like video synthesis~\cite{Gur2020HierarchicalPV}, structure analogy~\cite{Benaim2021StructuralAF} and inpainting~\cite{Cao2020GeneratorPF}.  
	 Although SinGAN has proved the feasibility of generating diverse images from a natural single image,  experiments show that it is still difficult to obtain plausible syntheses when the given image is complex, \eg, image with large objects. This drawback also exists in previous related works \cite{Hinz2020ImprovedTF,Gur2020HierarchicalPV,Chen2021MOGANMG,Granot2021DropTG,Sushko2021OneShotGL}, which almost follow the paradigm of SinGAN to build the pyramid networks for internal learning \cite{shocher2018zero}. 
	 
	 \par
\begin{figure*}[t]
	\centering
	\includegraphics[width=0.9\linewidth]{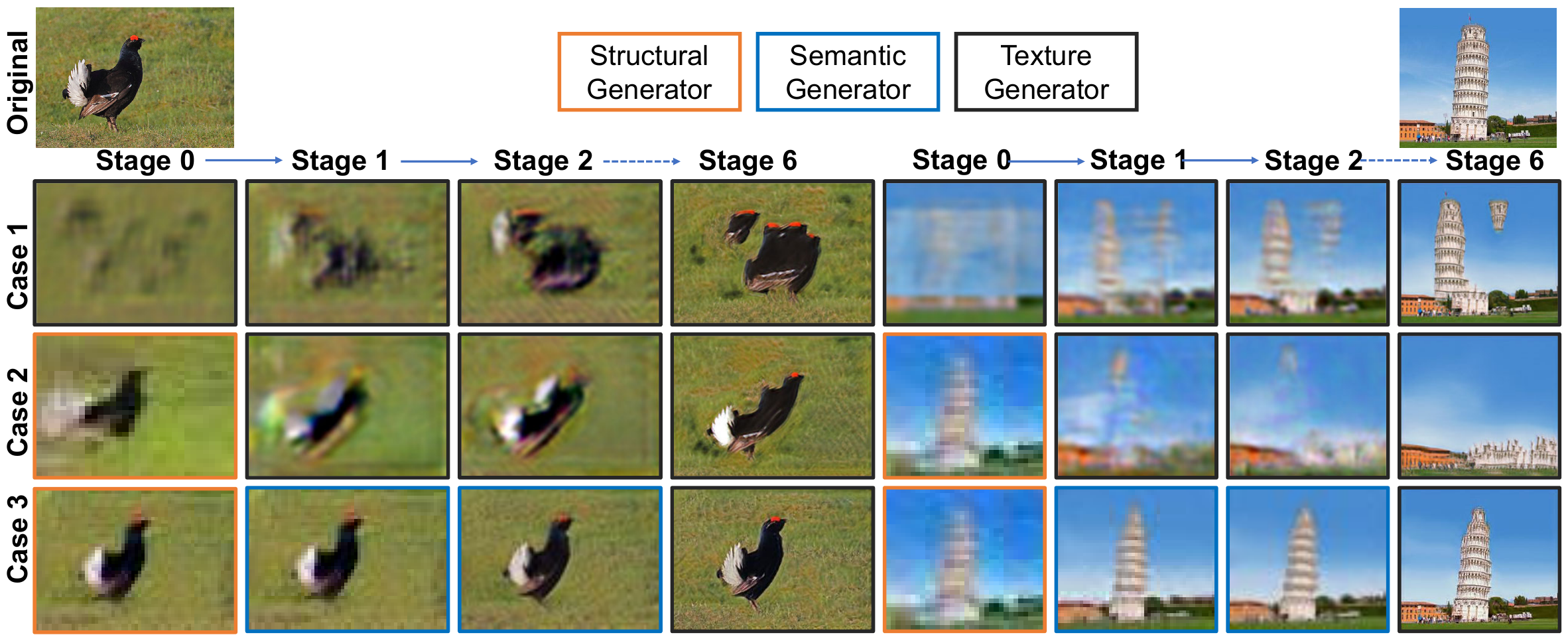}
	\caption{Each row illustrates the outputs from 7-stage single image generative models comprising generators in legend (Sec. \ref{Approach} for details), and the border color of each output implies its source generator. All pictures have been upsampled to the same size as the original. Note that Case 1 and Case 3 stand for SinGAN \cite{Shaham2019SinGANLA} and our ExSinGAN respectively.}
	\label{fig:examples}
\end{figure*}

 In this paper,  we focus on improving the generative capability of SinGAN on arbitrary natural images, especially for those images including objects or complex scenes. Firstly, we revisit the core idea of SinGAN,  namely the progressive learning strategy of pyramid networks. The progressive learning strategy \cite{Denton2015DeepGI,karras2017progressive} training the pyramid networks stage by stage, has played an important role in stabilizing the performance of GANs.
	 Specifically, for the large-scale progressively training model, the generator at the bottom learns to synthesize diverse but reasonable layouts of low resolution. In the subsequent stages, generators learn to add more semantic and texture details to the layouts. 
	 However, it is hard to obtain effective semantic and structural supervision from a single image. We observe that the bottom stage of SinGAN~\cite{Shaham2019SinGANLA} generally synthesizes disorder layouts (Case 1 in Fig. \ref{fig:examples}), and the details replenished in the later stages are likely to be uncontrollable and meaningless. This phenomenon illustrates the difference between SinGAN and the large-scale progressive training model in nature, where each stage of SinGAN plays an unclear or even negative role in the generation. The entire model is more like a magic black box we can never predict how it will run and behave. 
	\par Our insight is that, both the reliability and generative capability of SinGAN can be improved by introducing additional structural and semantic supervision.  In this paper, we propose a hierarchical framework for the single image generation problem and construct an \textit{explainable single image generative model} called ExSinGAN. Specifically, ExSinGAN is learned from both internal patches and external priors of the given image via three modular GANs.  Firstly, the structural GAN is devoted to synthesizing coarse but reasonable layout with the supervision of structural prior knowledge from GAN inversion~\cite{pan2020exploiting}. Then, the semantic GAN  is dedicated to injecting semantic details into the coarse layout with the supervision of a pre-trained classifier \cite{Johnson2016Perceptual}. Finally, the texture GAN replenishes fine texture details to the previous output.  To our best knowledge, ExSinGAN is the first single image generative model considering both internal information~\cite{Shocher2018ZeroShotSU, Shaham2019SinGANLA} and external generative priors~\cite{pan2020exploiting}.  Through experiments (Sec. \ref{Experiments}) we demonstrate that ExSinGAN can be adapted to a variety of images including texture, objects and scenes. Moreover, ExSinGAN also has competitive generalization capabilities on some image manipulation tasks like editing, harmonization and paint-to-image.
	
	\section{Related Work}\label{Related}

	\noindent\textbf{Internal Learning Generative Model.} 
	Deep internal learning methods, which are trained by the internal information of a single image, have taken off and caught popular attention  since ZSSR~\cite{Shocher2018ZeroShotSU} and DIP~\cite{ulyanov2018deep}. InGAN \cite{Shocher2019InGANCA} is the first internal learning generative model, which requires the pre-defined geometric transformation inputs to generate diverse images.  SinGAN~\cite{Shaham2019SinGANLA} is the first unconditional generative model, which comprises a pyramid of fully convolutional GANs to learn the internal patch distribution of the given image. 
SinGAN adopts the cascaded training strategy \cite{Denton2015DeepGI}: the bottom generator synthesizes the layout, and the later generators learn residual details to enlarge the layout. When the current generator is being trained, all lower stages generator are kept fixed. Some works have modified SinGAN from different perspectives. ConSinGAN~\cite{Hinz2020ImprovedTF} adopts the training strategy of PGGAN \cite{karras2017progressive} to improve the generative quality of SinGAN, and reduce the stages of the model to realize faster training. HP-VAEGAN~\cite{Gur2020HierarchicalPV} introduces patch-based VAE \cite{kingma2013auto} to make the generation more diverse.
	MOGAN, which~\cite{Chen2021MOGANMG} follows the multi-stage architecture of SinGAN, firstly synthesizes the interesting region and the rest region of the image respectively, then merges them into an integrated image.  GPNN~\cite{Granot2021DropTG} borrows the idea from the classical method PatchMatch~\cite{Barnes2009PatchMatchAR}, replacing each stage GAN with a non-training patch match module to generate a new image quickly. One-Shot GAN~\cite{Sushko2021OneShotGL} is a one-stage model containing multiple discriminators to learn different features of the given image.

	\noindent\textbf{Generative Priors and Model Inversion. }
	Different from the internal learning method, model inversion utilizes the pre-trained deep learning models as the external generative priors to conduct image synthesis. The most used pre-trained models include the classifiers \cite{Wang2021IMAGINEIS} and GANs~\cite{Xia2021GANIA}. In this paper, we focus on the most popular way, GAN inversion. GAN inversion aims at finding a code in the latent space of a pre-trained GAN that best reconstructs the given image. Once the optimal latent code is found, new images can be synthesized by manipulating the latent code. The methods of GAN inversion can be divided into four categories. The first one based on gradient optimization minimizes the loss function directly~\cite{zhu2016generative, brock2016neural,lipton2017precise,yeh2017semantic,shah2018solving,creswell2018inverting, raj2019gan}. The second one trains an encoder coupling with the generator to find the latent code indirectly~\cite{donahue2016adversarial, Donahue2019LargeSA}. The third one combining the above two methods together~\cite{zhu2016generative, Bau2019GANDV,ORegan2001ASA}, initializes the latent code via a coupled encoder, then optimizes the latent code using gradient optimization method.  The last one is joint optimization~\cite{bau2020semantic,pan2020exploiting}, which not only optimizes the latent code but also fine-tunes the parameters of the generator simultaneously. The representative model DGP~\cite{pan2020exploiting} of joint optimization fine-tunes the generator of BigGAN~\cite{brock2018large} pre-trained on ImageNet~\cite{deng2009imagenet} in a progressive manner, giving rise to more precise and faithful reconstruction for natural images.

	\begin{figure*}[t]
		\centering
		
		\includegraphics[width=0.9\linewidth]{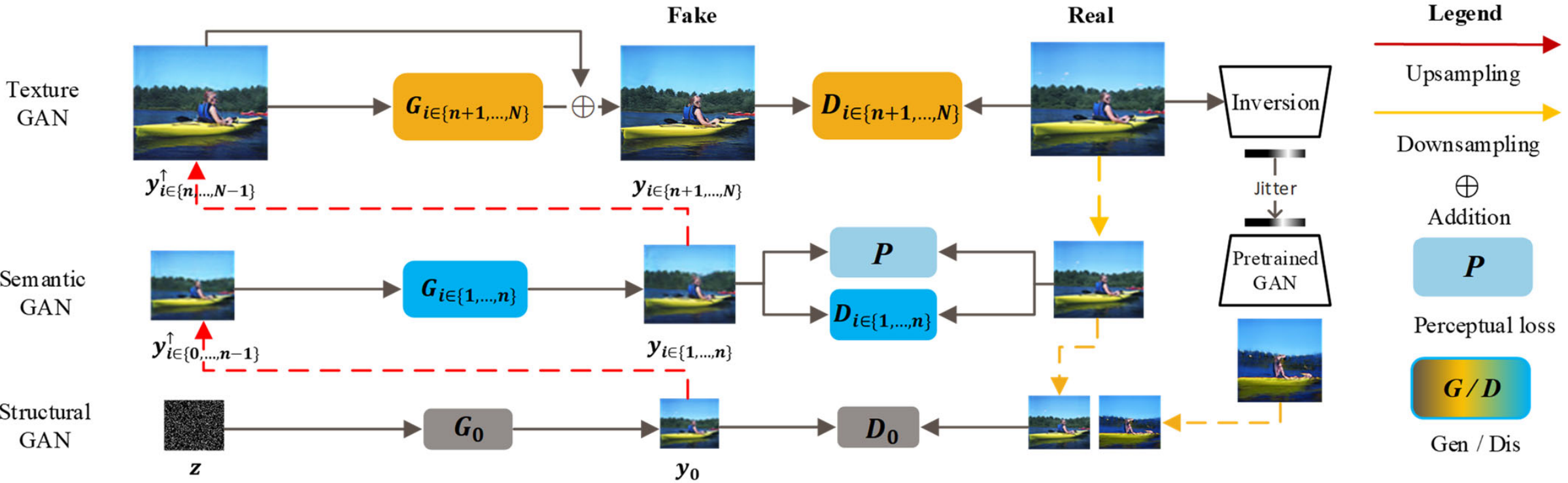}
		\caption{The architecture of ExSinGAN, which is composed of three modular GANs, the structural GAN for generating the coarse layout, semantic GAN for constraining the semantics of synthesis, the texture GAN for making the synthesis exquisite. Except for structural GAN, both semantic GAN and texture GAN are multi-stage. The GAN inversion provides a structural prior to train structural GAN.}
		\label{fig:exsingan}
	\end{figure*}
	

	\section{Methodology}\label{Approach}
	\noindent\textbf{Preliminaries.}  
		Considering a single image $x \in \mathbb{R}^{H\times W} $ has size $H\times W$,  a single image generative model formulates the conditional distribution $ p(y|x) $, which we can sample from to get the synthesis $ y \in \mathbb{R}^{H\times W} $. For convenience, we use $x^{i} \in \mathbb{R}^{H_i\times W_i}$ to denote the downsampled version of $x$, where the size of  $ x^{i} $ increases with the stage $i$ varying from $ 0 $ to $ N $ and $ x = x_N $.  Instead of learning the distribution $ p(y|x) $, the progressive training of SinGAN aims at learning
		\begin{equation} \label{multi-scale eq}
			\setlength{\abovedisplayskip}{3pt} 
			\setlength{\belowdisplayskip}{3pt}
			\begin{split}
				p(y^0, y^1, \dots,  y^N | x^0, x^1, \dots,  x^N )
				& \approx p(y^0 | x^0)\mathop{\Pi}^{N}_{i=1}p(y^{i} | x^{i}, y^{i-1}),
			\end{split} 
		\end{equation}
	  where  $y^i $ can be sampled successively to get the final synthesis $y^N$. 		
	 We make an assumption that image $x$ can be  decomposed into $ \left\lbrace  x_{str}, x_{sem}, x_{tex}\right\rbrace $ called the \textit{triplet representation}, where $ x_{str}, x_{sem}$ and $x_{tex}$ represent the structure, semantics and texture of $ x $ respectively. 
	
	\par We firstly point that such representation exists and  $  x_{str}, x_{sem}, x_{tex} $ can be well-defined on specific manifolds. 	
	Firstly, Considering a large-scale pre-trained generator $G_{pre}$ and $z^*$ subjected to $x=G_{pre}(z^*)$, we formulate $x_{str}:={G_{pre}(z^*)}^0$. The superscript `$0$' represents the downsampling operator as defined in Eq. \eqref{multi-scale eq}, which aims to compress the texture and keep the layout of $x$ effectively. Note that even though  $x_{str}$ and $x^0$ are equal, $x_{str}$ belongs to the new manifold defined by $G_{pre}(z), z \in $ latent space. Unlike that disturbance in pixel space usually brings noise to image, the disturbance to latent code $z^*$ likely changes the structure of $x$ reasonably, where the neighbors of $x_{str}$ have similar structure to $x_{str}$.   Then, the semantic feature $x_{sem}$ can be extracted using pre-trained convolutional networks $\phi$ such as VGG19 \cite{Simonyan2015VeryDC}, \textit{i.e.}, $x_{sem}=\phi(x)$. Finally, the texture feature $x_{tex}$ can be interpreted as the internal patches of $x$. Thus, we have determined the form of the triplet. 

		\noindent\textbf{Proposed Method.}  
	It is not difficult to find that SinGAN just constrains the distance between $y_{tex} $ and  $x_{tex}$
	by patch discriminators,  \textit{i.e.}, maximizing $p(y_{tex}^{0}|x_{tex}^{0}) \mathop{\Pi}^{N}_{i=1}p(y_{tex}^{i} | x_{tex}^{i}, y^{i-1})$, 
	hence SinGAN hardly generates images with reasonable structure and semantics.   According to triplet representation, we hope to maximize Eq. (\ref{eq:2}) to make all structure, semantics and texture reasonable. Therefore, we introduce the strengthening form of Eq. (\ref{multi-scale eq})
	\begin{equation}\label{eq:2}
		\setlength{\abovedisplayskip}{3pt} 
		\setlength{\belowdisplayskip}{3pt}
		\begin{split}		
			p(y_{str}|x_{str}) p(y_{tex}^{0}|x_{tex}^{0}) \mathop{\Pi}^{n}_{i=1}(p(y^{i}_{sem}|x^{i}_{sem}, y^{i-1})p(y_{tex}^{i} | x_{tex}^{i}, y^{i-1})) \mathop{\Pi}^{N}_{i=n+1}p(y_{tex}^{i} | x_{tex}^{i}, y^{i-1}).
		\end{split}   	
	\end{equation}
	Eq. \eqref{eq:2} consists of three parts, the fist part $p(y_{str}|x_{str})$$p(y_{tex}^{0}|x_{tex}^{0})$ makes the texture and structure of syntheses similar to the origin. The second part $\mathop{\Pi}^{n}_{i=1}(p(y^{i}_{sem}|x^{i}_{sem}, y^{i-1})p(y_{tex}^{i} | x_{tex}^{i}, y^{i-1}))$ makes the texture and semantics of syntheses similar to the origin at the intermediate stages. and the third part $\mathop{\Pi}^{N}_{i=n+1}p(y_{tex}^{i} | x_{tex}^{i}, y^{i-1})$ only makes the texture of syntheses similar to the origin in high resolution.
	Fig. \ref{fig:exsingan} shows the overview of ExSinGAN, where the network comprises three modular GANs, respectively structural GAN, semantic GAN, and texture GAN aiming to model three parts in Eq.(\ref{eq:2}) successively.  
	In the following section, we simply denote each generator and discriminator of ExSinGAN as $G_i$ and $D_i$, $i \in \{0,\dots, N\}$.\\
	\noindent\textbf{Structural GAN.} The structural GAN implicitly learns $p(y_{tex}^{0}|x_{tex}^{0})p(y_{str}|x_{str})$ from the image $x$, where the generator $G_0$ synthesizes images of reasonable layouts and high quality of texture.  We firstly adopt GAN inversion method DGP~\cite{pan2020exploiting} to find  $z^*$ in the latent space of BigGAN \cite{brock2018large}, then disturb $z^*$ via Gaussian noise to generate dataset $Data =  \{G_{pre}(z^*+\Delta z_i)^0| i = 1,\dots,m \}$, where $\Delta z_i$ is a random noise and  $m$ is the number of syntheses. Moreover,  because the texture and color of syntheses in $Data$ differ greatly from the original image, we insert some copies of $x^0$ into $Data$ to suppress these artifacts. $ D_0 $ is designed as a fully-connected discriminator to learn the texture and structure together, which is different from that  SinGAN~\cite{Shaham2019SinGANLA} uses a patch discriminator. WGAN-gp loss~\cite{gulrajani2017improved} 
	\begin{equation}\label{3}
		\setlength{\abovedisplayskip}{3pt} 
		\setlength{\belowdisplayskip}{3pt}
		\mathop{\rm{min}}_{G_0}\mathop{\rm{max}}_{D_0} \mathcal{L}_{adv}(G_0, D_0) 
	\end{equation}
	is used as the adversarial loss to keep training stable. Sinusoidal positional encoding is concatenated with the input noise~\cite{Devlin2019BERTPO} to enhance the spatial inductive bias of  $ G_0 $~\cite{Xu2020PositionalEA}. Once the structural GAN is trained, $ G_0 $ can generate low-resolution layouts of higher quality compared with SinGAN (stage 0 in Fig. \ref{fig:examples}).
	
	\noindent\textbf{Semantic GAN.} According to $\mathop{\Pi}^{n}_{i=1}(p(y^{i}_{sem}|y^{i}_{sem})p(y_{tex}^{i} | x_{tex}^{i}, y^{i-1}))$, semantic GAN amis at adding more texture and semantic details to the layout from $G_0$. For the semantic details, we adopt the perceptual loss~\cite{Johnson2016Perceptual}
	\begin{equation}
	    \mathcal{L}_{p}(x^i, G_i(y^{i-1\uparrow})) = \Vert   \phi\left(x\right)-\phi(G_i( y^{i-1\uparrow}))\Vert
	\end{equation}
	to enlarge $x^0$ meaningfully, where "$ \uparrow $" means bicubic interpolation, and $G_i$ is a fully convolutional network. Here we do not use residual learning like SinGAN, because it will cause serious artifacts in practice. Patch discriminator $D_i$ is used for patch adversarial learning, and the reconstruction loss $\mathcal{L}_{\mathrm{rec}}(G_i) =\Vert   G_i(x^{{i-1}^{\uparrow}})-x^i\Vert^2$ aims to stabilize the training. Denoting $\mathcal{L}_{p}(x^i, G_i(y^{i-1\uparrow})) $ as $ \mathcal{L}_{p}(G_i) $,  the total training loss is  
	\begin{equation}\label{eq:semantic loss}
		\setlength{\abovedisplayskip}{3pt} 
		\setlength{\belowdisplayskip}{3pt}
		\mathop{\rm{min}}_{G_i} \mathop{\rm{max}}_{D_i} \mathcal{L}_{adv}(G_i, D_i) + \alpha_1 \mathcal{L}_{rec}(G_i) + \lambda \mathcal{L}_{p}(G_i).
	\end{equation}
 	For arbitrary $ y^{0} $ from $ G_0 $,  the output $ y^{n} $ has similar layout to $ y^0 $ and close semantics to $x$, and without semantic generator the quality will be poor (Case 2 and Case 3 in Fig. \ref{fig:examples}). 
	 The reason we design multi-stage semantic GANs is that,  we observed that only one semantic GAN layer cannot provide effective semantic supervision. However, existing research~\cite{Sajjadi2017EnhanceNetSI,zhou2018non} also proves that the perceptual loss is helpful for better visual quality on image restoration tasks, nevertheless produces artifacts into the structures and causes color distortion. Hence a proper $n$ is important and will be discussed it in Sec. \ref{Experiments}. It is worth noting that semantic constraint has been mentioned in recent work IMAGINE \cite{Wang2021IMAGINEIS}, where a semantic constraint is used to guide the category information of syntheses. Whereas in our approach semantic constraint aims to replenish specific semantic details on the layout. 
	
	\noindent\textbf{Texture GAN.} The last term $\mathop{\Pi}^{N}_{i=n+1}p(y_{tex}^{i} | x_{tex}^{i}, y^{i-1})$ indicates that adding  texture details to $ y^n $ based on the reference $ x $.  We follow SinGAN to adopt the the multi-stage architecture with the residual block that is widely used in super-resolution models~\cite{kim2016accurate}
	\begin{equation}
		\setlength{\abovedisplayskip}{3pt} 
		\setlength{\belowdisplayskip}{3pt}
		y^{i} = G_i(y^{i-1{\uparrow}}) + y^{i-1{\uparrow}},\ i=n+1,\dots,N.
	\end{equation}
	The training loss contains the adversarial loss and reconstruction loss,
	\begin{equation}\label{key}
		\setlength{\abovedisplayskip}{3pt} 
		\setlength{\belowdisplayskip}{3pt}
		\mathop{\rm{min}}_{G_i} \mathop{\rm{max}}_{D_i} \mathcal{L}_{adv}(G_i, D_i) + \alpha_2 \mathcal{L}_{rec}(G_i).
	\end{equation}

\begin{figure}
		\centering
		\includegraphics[width=0.7\linewidth]{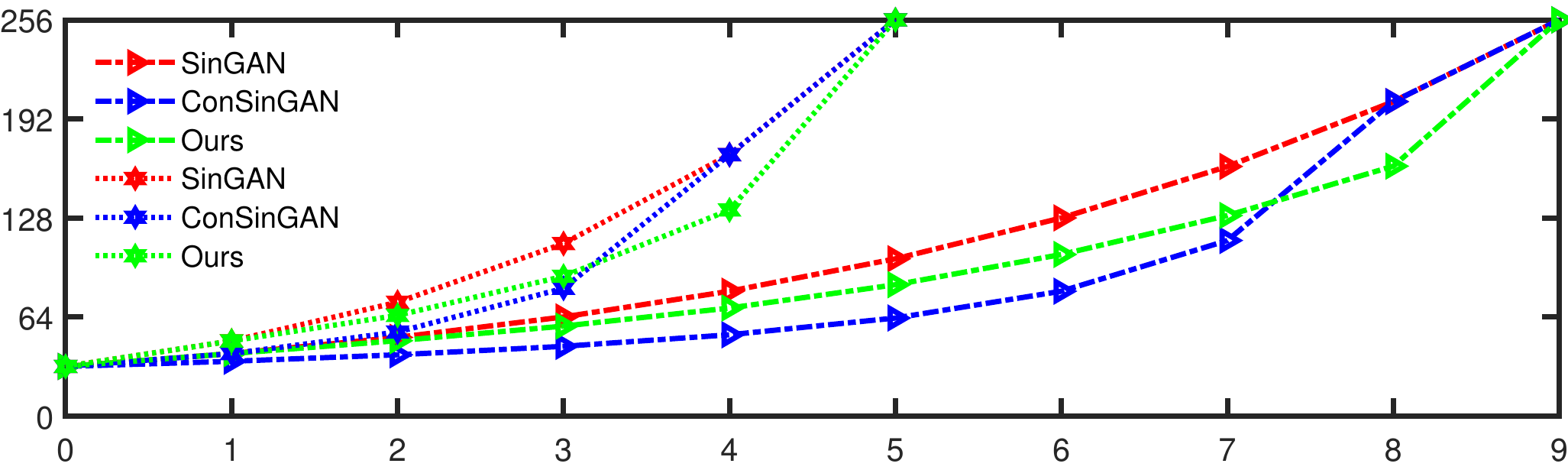}
		\caption{Comparison of different rescaling methods when $ N =5 $ and $ 9 $, the horizontal and vertical axes respectively represent stage and pixel.}
		\label{fig:taylor__}
	\end{figure}

	\noindent\textbf{Implementation.}\label{Implementation} To speed up the training, except that $ D_0 $ comprises three convolutional layers and one fully connected layer,  all other $ G_i $ and $ D_i $ are simply composed of five convolutional layers. Supposed the height $ H $ of $ x $ is longer than the width $ W $,  the longer side $ H_N $ of $ x^N$  is set to no more than $ 256 $ and the shorter side $W_0$ of $x^0$ is fixed to $32$. The rescaling method influences both quality of synthesis and training time.   See Fig. \ref{fig:taylor__}, SinGAN \cite{Shaham2019SinGANLA} takes the basic rescaling method $ H_s =  H_N \times r^{N-s} $,   where $ r $ is scalar factor, and $s$ is current stage. ConSinGAN \cite{Hinz2020ImprovedTF} takes a complicated rescaling method to focus on the lower stage. However, we find that these rescaling methods are  time-consuming and not robust (see Appendix for details).
	Hence we design a new rescaling method by analogous Taylor approximation of the basic rescaling method
	\begin{equation}\label{Taylor}
		\setlength{\abovedisplayskip}{3pt} 
		\setlength{\belowdisplayskip}{3pt}
		H_s =  H_0\times (1+st+\frac{s(s-1)(s-2)}{k}t^3), \ s = 0,\dots,N.
	\end{equation}  
	In above equation, $ t = \frac{1}{r} -1 $ and $ k $ is adjustable to control the curve shape. In this paper we set $ k=2 $, $ r $ is determined by $ N $. Fig. \ref{fig:taylor__} shows the slope of our rescaling method does not vanish at original point, and becomes larger when $ s $ approaches to $ N $,  which makes the model more concerned with the generation at intermediate stages and less time-consuming at top stages.
	\par For training structural GAN, we jitter the latent code by Gaussian noise with mean 0, and standard deviations varying from $0.1$ to $0.5$ to get 500 training data in total. Additional 150 copies of the original images are also injected into training data. 
	Existing works used the convolutional layers of pre-trained VGG-19~\cite{Simonyan2015VeryDC} for computing perceptual loss. We have tried the different convolutional layers of VGG-19 in advance and concluded that the combination of layers from \texttt{relu51} to \texttt{relu53} is the best choice for our model. 
	To take trade of training time and performance, we set $ N = 6 $, $ n = 3 $,  $ \lambda = 0.1 $, $ \alpha_1, \alpha_2 =10 $. For structural GAN, we set training epochs to 10k, learning rate to $ 10^{-4} $, batch size to $ 32 $ for both $ G_0 $ and $ D_0 $. For the semantic and texture GAN, we set training epochs to $ 2 $k, learning rate to $ 5 \times10^{-4} $ for $ G_i $ and $ D_i $, $ i = 1,\dots, N $. 

	\section{Experimental Results}\label{Experiments}
In this section, we qualitatively and quantitatively compare ExSinGAN with DGP \cite{pan2020exploiting}, SinGAN \cite{Shaham2019SinGANLA}, ConSinGAN\cite{Hinz2020ImprovedTF}, HP-VAEGAN \cite{Gur2020HierarchicalPV} in terms of image synthesis, editing, harmonization and paint-to-image tasks, then conduct ablation study to demonstrate that each module of ExSinGAN is significant. 
	
	\begin{figure}[t]
		\includegraphics[width= 1\linewidth]{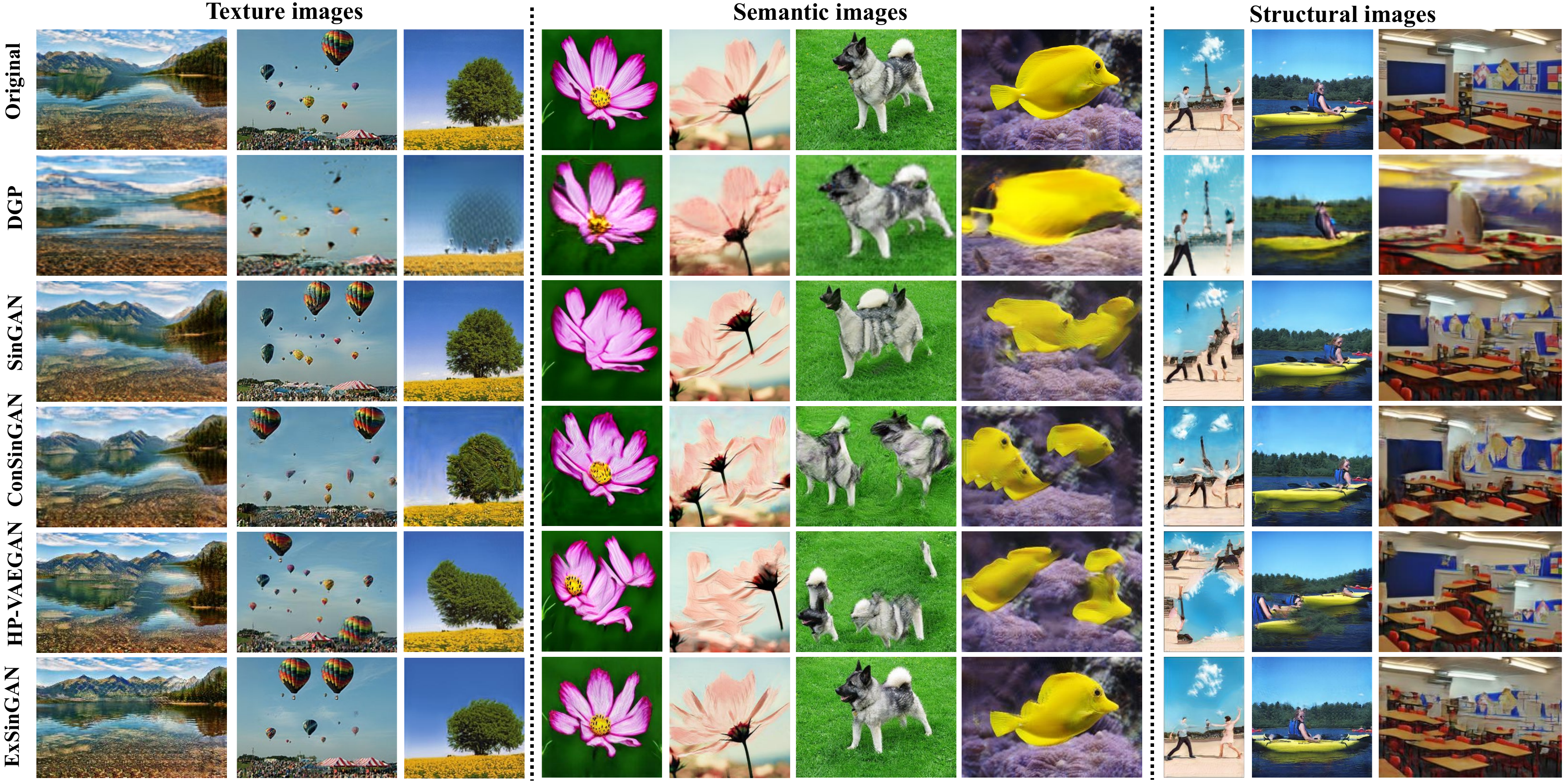}
		\caption{Comparison of DGP \cite{pan2020exploiting}, SinGAN \cite{Shaham2019SinGANLA}, ConSinGAN\cite{Hinz2020ImprovedTF}, HP-VAEGAN \cite{Gur2020HierarchicalPV} and ExSinGAN on image synthesis. }
		\label{fig:syntheses}
	\end{figure}
	
	\begin{figure}[t]
		\setlength{\belowcaptionskip}{-0.5cm} 
		\begin{center}
			
			\begin{minipage}{0.33\linewidth}
				\begin{minipage}{0.185\linewidth}	
					\centerline{\includegraphics[width=1\textwidth]{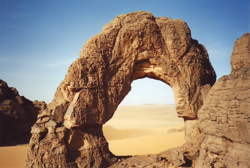}}
				\end{minipage}
				\begin{minipage}{0.185\linewidth}	
					\centerline{\includegraphics[width=1\textwidth]{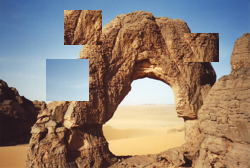}}
				\end{minipage}
				\begin{minipage}{0.185\linewidth}	
					\centerline{\includegraphics[width=1\textwidth]{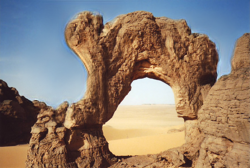}}
				\end{minipage}
				\begin{minipage}{0.185\linewidth}	
					\centerline{\includegraphics[width=1\textwidth]{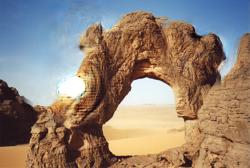}}
				\end{minipage}
				\begin{minipage}{0.185\linewidth}	
					\centerline{\includegraphics[width=1\textwidth]{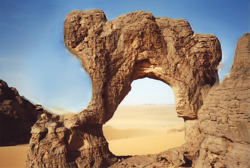}}
				\end{minipage}
				
				\begin{minipage}{0.185\linewidth}	
					\centerline{\includegraphics[width=1\textwidth]{fig/editing/stone}}
				\end{minipage}
				\begin{minipage}{0.185\linewidth}	
					\centerline{\includegraphics[width=1\textwidth]{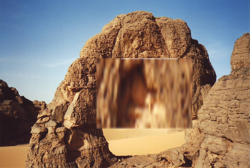}}
				\end{minipage}
				\begin{minipage}{0.185\linewidth}	
					\centerline{\includegraphics[width=1\textwidth]{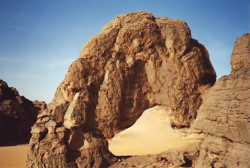}}
				\end{minipage}
				\begin{minipage}{0.185\linewidth}	
					\centerline{\includegraphics[width=1\textwidth]{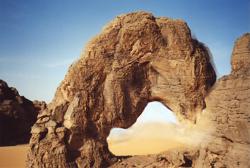}}
				\end{minipage}
				\begin{minipage}{0.185\linewidth}	
					\centerline{\includegraphics[width=1\textwidth]{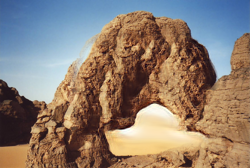}}
				\end{minipage}
				
				\begin{minipage}{0.185\linewidth}	
					\centerline{\includegraphics[width=1\textwidth]{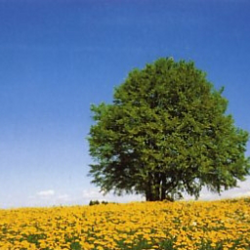}}
				\end{minipage}
				\begin{minipage}{0.185\linewidth}	
					\centerline{\includegraphics[width=1\textwidth]{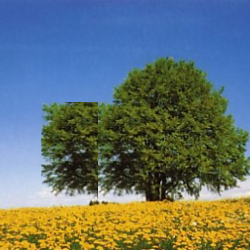}}
				\end{minipage}
				\begin{minipage}{0.185\linewidth}	
					\centerline{\includegraphics[width=1\textwidth]{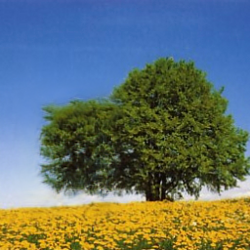}}
				\end{minipage}
				\begin{minipage}{0.185\linewidth}	
					\centerline{\includegraphics[width=1\textwidth]{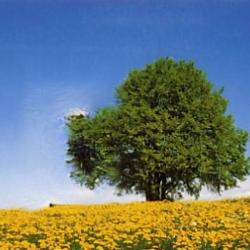}}
				\end{minipage}
				\begin{minipage}{0.185\linewidth}	
					\centerline{\includegraphics[width=1\textwidth]{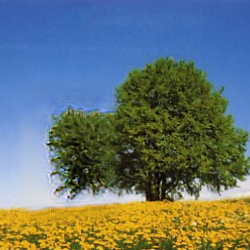}}
				\end{minipage}
				
				\begin{minipage}{0.185\linewidth}	
					\centerline{\includegraphics[width=1\textwidth]{fig/editing/tree}}
				\end{minipage}
				\begin{minipage}{0.185\linewidth}	
					\centerline{\includegraphics[width=1\textwidth]{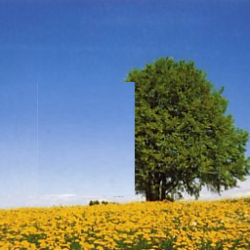}}
				\end{minipage}
				\begin{minipage}{0.185\linewidth}	
					\centerline{\includegraphics[width=1\textwidth]{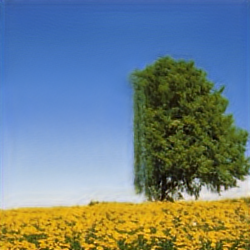}}
				\end{minipage}
				\begin{minipage}{0.185\linewidth}	
					\centerline{\includegraphics[width=1\textwidth]{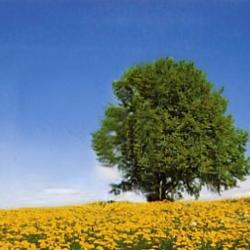}}
				\end{minipage}
				\begin{minipage}{0.185\linewidth}	
					\centerline{\includegraphics[width=1\textwidth]{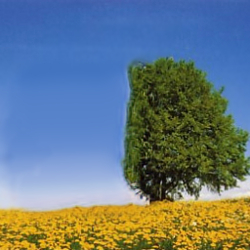}}
				\end{minipage}
				\label{editing}
			\end{minipage}
			\begin{minipage}{0.306\linewidth}
				\begin{minipage}{0.185\linewidth}	
					\centerline{\includegraphics[width=1\textwidth]{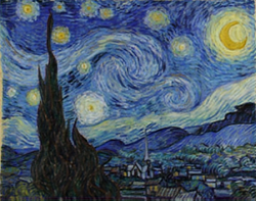}}
				\end{minipage}
				\begin{minipage}{0.185\linewidth}	
					\centerline{\includegraphics[width=1\textwidth]{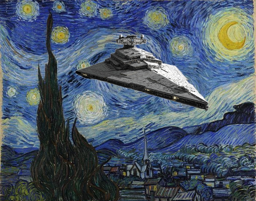}}
				\end{minipage}
				\begin{minipage}{0.185\linewidth}	
					\centerline{\includegraphics[width=1\textwidth]{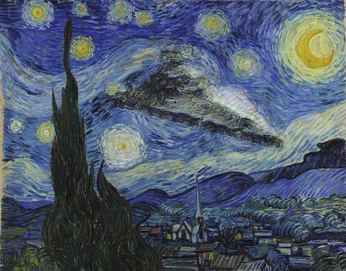}}
				\end{minipage}
				\begin{minipage}{0.185\linewidth}	
					\centerline{\includegraphics[width=1\textwidth]{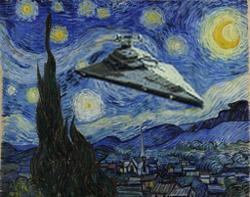}}
				\end{minipage}
				\begin{minipage}{0.185\linewidth}	
					\centerline{\includegraphics[width=1\textwidth]{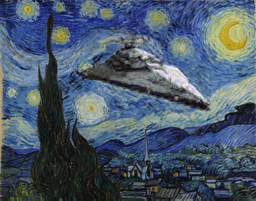}}
				\end{minipage}
				
				\begin{minipage}{0.185\linewidth}	
					\centerline{\includegraphics[width=1\textwidth]{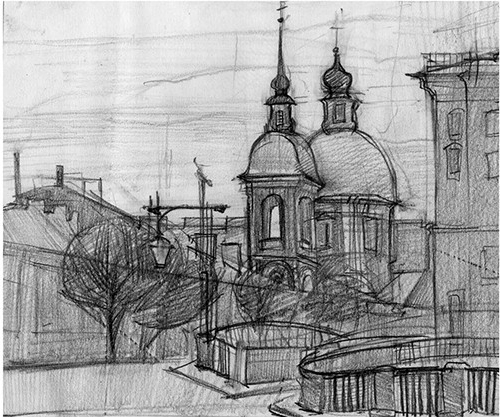}}
				\end{minipage}
				\begin{minipage}{0.185\linewidth}	
					\centerline{\includegraphics[width=1\textwidth]{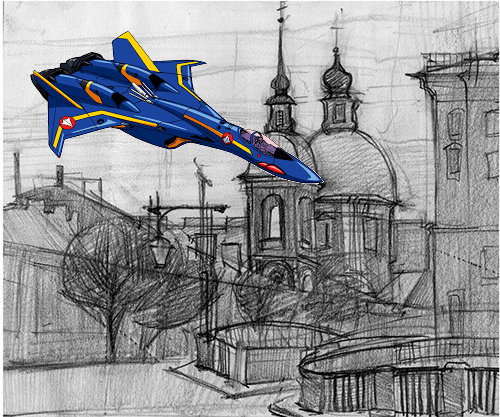}}
				\end{minipage}
				\begin{minipage}{0.185\linewidth}	
					\centerline{\includegraphics[width=1\textwidth]{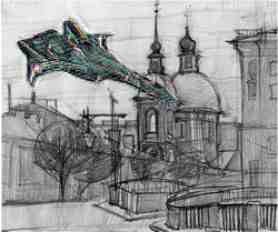}}
				\end{minipage}
				\begin{minipage}{0.185\linewidth}	
					\centerline{\includegraphics[width=1\textwidth]{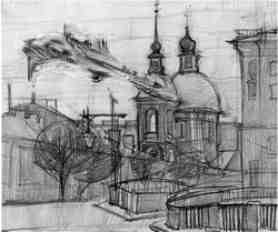}}
				\end{minipage}
				\begin{minipage}{0.185\linewidth}	
					\centerline{\includegraphics[width=1\textwidth]{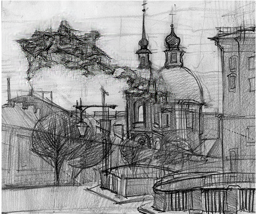}}
				\end{minipage}
				
				\begin{minipage}{0.185\linewidth}	
					\centerline{\includegraphics[width=1\textwidth]{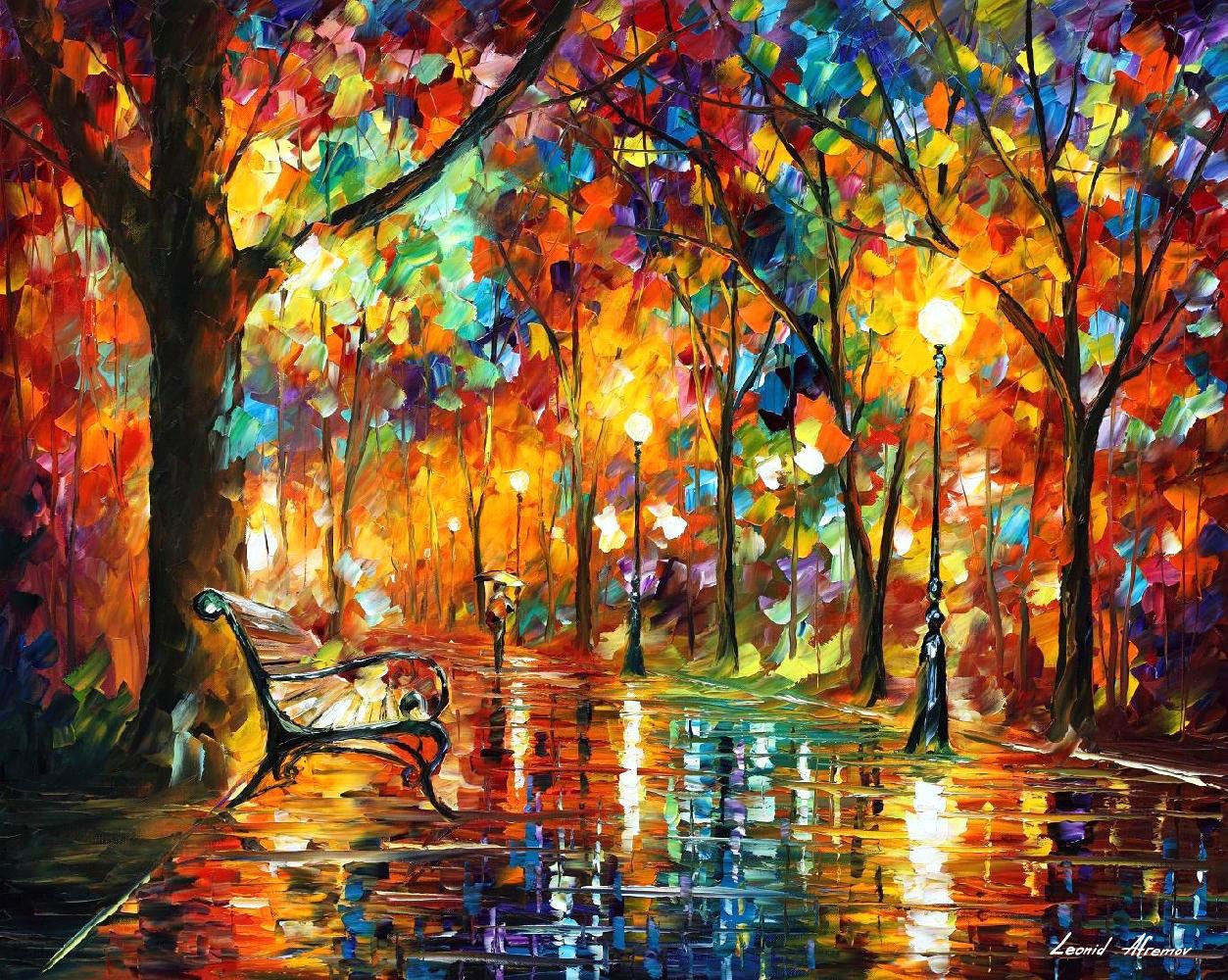}}
				\end{minipage}
				\begin{minipage}{0.185\linewidth}	
					\centerline{\includegraphics[width=1\textwidth]{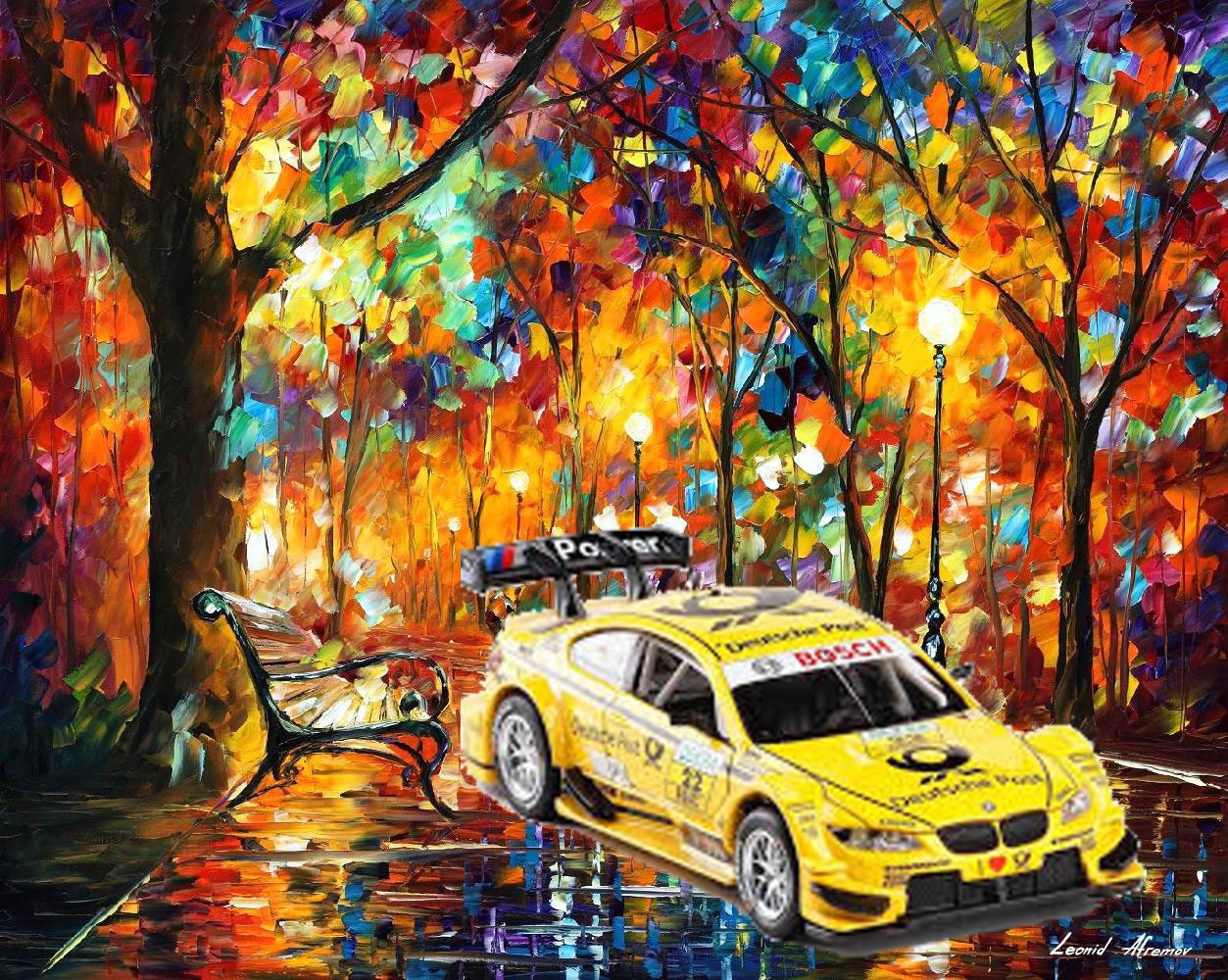}}
				\end{minipage}
				\begin{minipage}{0.185\linewidth}	
					\centerline{\includegraphics[width=1\textwidth]{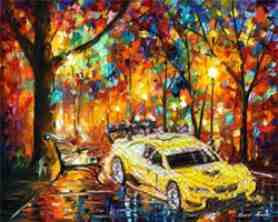}}
				\end{minipage}
				\begin{minipage}{0.185\linewidth}	
					\centerline{\includegraphics[width=1\textwidth]{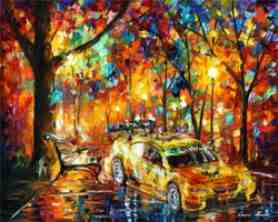}}
				\end{minipage}
				\begin{minipage}{0.185\linewidth}	
					\centerline{\includegraphics[width=1\textwidth]{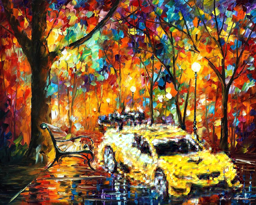}}
				\end{minipage}

				\begin{minipage}{0.185\linewidth}	
					\centerline{\includegraphics[width=1\textwidth]{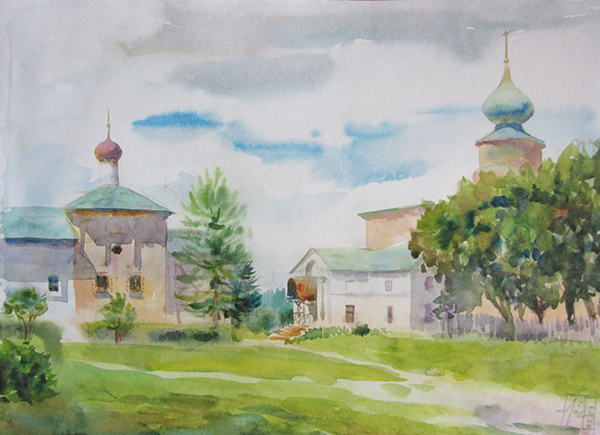}}
				\end{minipage}
				\begin{minipage}{0.185\linewidth}	
					\centerline{\includegraphics[width=1\textwidth]{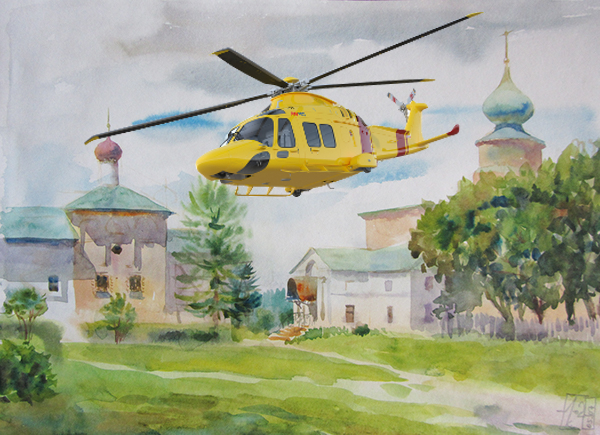}}
				\end{minipage}
				\begin{minipage}{0.185\linewidth}	
					\centerline{\includegraphics[width=1\textwidth]{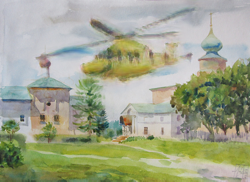}}
				\end{minipage}
				\begin{minipage}{0.185\linewidth}	
					\centerline{\includegraphics[width=1\textwidth]{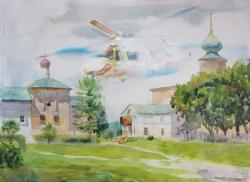}}
				\end{minipage}
				\begin{minipage}{0.185\linewidth}	
					\centerline{\includegraphics[width=1\textwidth]{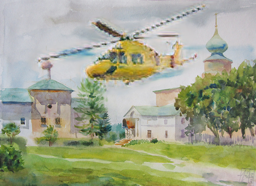}}
				\end{minipage}
				\label{harmonization}
			\end{minipage}
			\begin{minipage}{0.3328\linewidth}
				
				\begin{minipage}{0.235\linewidth}
					\centerline{\includegraphics[width=1\textwidth]{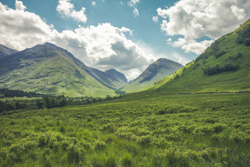}}
				\end{minipage}
				\begin{minipage}{0.235\linewidth}
					\centerline{\includegraphics[width=1\textwidth]{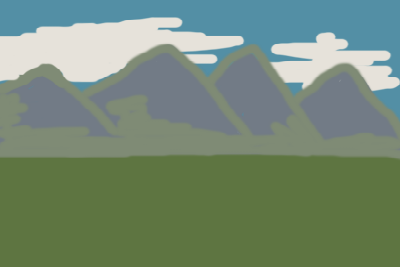}}
				\end{minipage}
				\begin{minipage}{0.235\linewidth}
					\centerline{\includegraphics[width=1\textwidth]{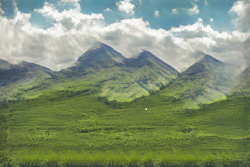}}
				\end{minipage}
				\begin{minipage}{0.235\linewidth}
					\centerline{\includegraphics[width=1\textwidth]{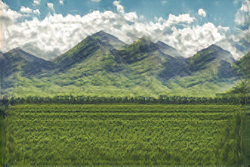}}
				\end{minipage}
				
				\begin{minipage}{0.235\linewidth}
					\centerline{\includegraphics[width=1\textwidth]{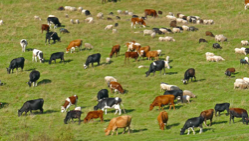}}
				\end{minipage}
				\begin{minipage}{0.235\linewidth}
					\centerline{\includegraphics[width=1\textwidth]{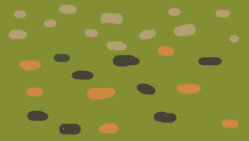}}
				\end{minipage}
				\begin{minipage}{0.235\linewidth}
					\centerline{\includegraphics[width=1\textwidth]{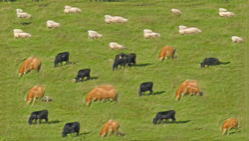}}
				\end{minipage}
				\begin{minipage}{0.235\linewidth}
					\centerline{\includegraphics[width=1\textwidth]{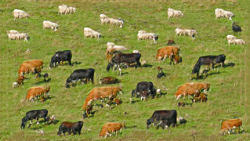}}
				\end{minipage}
				
				\begin{minipage}{0.235\linewidth}
					\centerline{\includegraphics[width=1\textwidth]{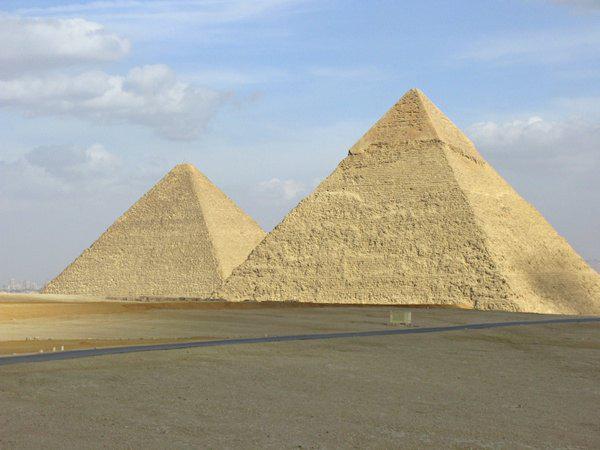}}
				\end{minipage}
				\begin{minipage}{0.235\linewidth}
					\centerline{\includegraphics[width=1\textwidth]{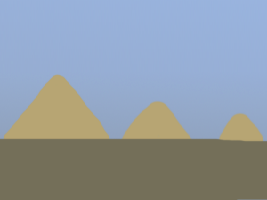}}
				\end{minipage}
				\begin{minipage}{0.235\linewidth}
					\centerline{\includegraphics[width=1\textwidth]{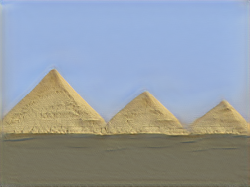}}
				\end{minipage}
				\begin{minipage}{0.235\linewidth}
					\centerline{\includegraphics[width=1\textwidth]{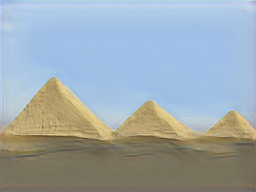}}
				\end{minipage}
				
				\begin{minipage}{0.235\linewidth}
					\centerline{\includegraphics[width=1\textwidth]{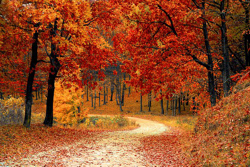}}
				\end{minipage}
				\begin{minipage}{0.235\linewidth}
					\centerline{\includegraphics[width=1\textwidth]{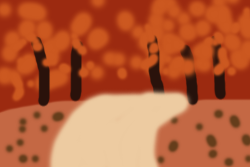}}
				\end{minipage}
				\begin{minipage}{0.235\linewidth}
					\centerline{\includegraphics[width=1\textwidth]{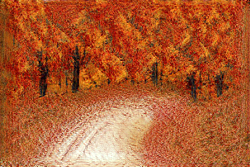}}
				\end{minipage}
				\begin{minipage}{0.235\linewidth}
					\centerline{\includegraphics[width=1\textwidth]{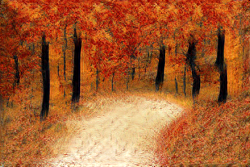}}
				\end{minipage}
				\label{paint}

			\end{minipage}	
			
			\begin{minipage}{0.33\linewidth}
				\centerline{(a). Editing}
			\end{minipage}
			\begin{minipage}{0.306\linewidth}
				\centerline{(b). Harmonization}
			\end{minipage}
			\begin{minipage}{0.3328\linewidth}
				\centerline{(c). Paint-to-image}
			\end{minipage}	
			
		\end{center}
		\vspace{-1em}
		\caption{Comparison of SinGAN \cite{Shaham2019SinGANLA}, ConSinGAN\cite{Hinz2020ImprovedTF}, ExSinGAN on editing, harmonization and paint-to-image. In (a) and (b), the five columns respectively are original, naive, SinGAN, ConSinGAN, ExSinGAN. In (c), the four columns are original, painting, SinGAN, ExSinGAN.}
		\label{syntheses}
		
	\end{figure}
	
	\subsection{Image Synthesis}
	\textbf{Qualitative Evaluation.}  We conduct experiments on various types of images for comprehensive comparison (Fig. \ref{fig:syntheses}), including texture image with rich texture but the simple structure, semantic image containing a single complicated object, and structural image comprising a complex scene with several objects. 
	
	Firstly, the generation of texture images primarily tests the ability to synthesize texture.  Although syntheses of DGP have unpromising texture, ExSinGAN can obtain exquisite results rivaling SinGAN by revising these unsatisfactory images. Note that the syntheses of ConSinGAN have artifacts and coarse texture, and this phenomenon persists when we increase its stages. 
	Secondly, free deformation of the structure and repetitive patches are no longer permitted in semantic images, which further tests the generative ability of models. We first illustrate the erratic generative ability of previous models with two flowers of similar structure in semantic images of Fig. \ref{fig:syntheses}. For the first flower, SinGAN and ConSinGAN are confused with the relation of petals and ignore the flower stalk, but the overall layouts of syntheses are plausible. However, they are failed to synthesize the second flower, where the syntheses are of unreasonable repetitive patches petals. In fact, through experiments we find that SinGAN and ConSinGAN are sensitive to color change very much.  Note that the color of the first flower is distinct from the background color, 
hence they can perceive the existence of the petals rather than the stalk. Similarly, they can perceive the existence of the stalk rather than the petals for the second flower. ExSinGAN not only successfully synthesizes the petals but also generates the stalk beneath the petals. 
For the dog image and fish image, except that syntheses of DGP have obvious texture defects, the other three generative methods are completely failed in vision. This is similar to case 1 shown in Fig, \ref{fig:examples}. Due to the lack of structural and semantic constraints, they can only generate repeated patches, while the syntheses of ExSinGAN can also maintain reasonable structure and high-quality texture. Thirdly, for the structural images, despite the fact DGP performs badly in texture, ExSinGAN just properly utilizes its merit of generating plausible layouts, making reasonable variations in images. Visually, ExSinGAN has made satisfying structural changes rather than simply distorting the image. For example, for the dancing image synthesized by ExSinGAN, although the tower in the background disappears, the structure of the image is not polluted like the syntheses of SinGAN. As can be seen from the boat and classroom images, ExSinGAN can produce stronger changes in the layout, proving that ExSinGAN is not overfitting to the given image.

	\begin{table*}[]
		\centering
		\setlength{\abovecaptionskip}{0.5em}
		 \resizebox{\textwidth}{20mm}
		{
			\begin{tabular}{@{}l|cc|cc|cc|c@{}}
				\toprule
				\textbf{Model}  & \multicolumn{2}{c|}{\textit{\textbf{Places50}}} & \multicolumn{2}{c|}{\textit{\textbf{LSUN50}}} & \multicolumn{2}{c|}{\textit{\textbf{ImageNet50}}} & \\
				\hline
				& SIFID$\downarrow$    & SSIM$\downarrow$    & SIFID$\downarrow$    & SSIM$\downarrow$  & SIFID$\downarrow$   &  SSIM$\downarrow$ & Training time  \\
				\hline
				DGP\cite{pan2020exploiting} & 0.64 &{0.18} &0.75 &{0.13} &1.65&{0.17} &  2 min\\ 
				SinGAN\cite{Shaham2019SinGANLA}     & 0.09      & 0.37   & 0.23      & 0.33  & 0.60      & 0.34  & 80 min\\
				
				ConSinGAN\cite{Hinz2020ImprovedTF}              & 0.06       & 0.25   & 0.11    & 0.26  & 0.56   & 0.23  & 25 min\\
				
				HP-VAEGAN\cite{Gur2020HierarchicalPV}               & 0.30    & 0.19   & 0.32   & 0.16  &{0.60}   & 0.16  & 63 min \\
				
				ExSinGAN               & 0.10     & 0.34   & {0.11}   & 0.35  & {0.45}   & 0.34 & 40 min\\
				
				\hline
				Te.7 & 0.24    & 0.19  & 0.53   & 0.14    & 0.51   & 0.19 & -\\
				St.1 Te.6   & 0.17    & 0.32   & 0.17     & 0.31  & 0.50      & 0.27& - \\
				
				St.1 Se.1 Te.5 &0.15 &0.34 & 0.15&0.34 &0.54&0.34 & -\\
				St.1 Se.3 Te.3  & 0.10      & 0.34   & 0.11   & 0.35  & 0.45     & 0.34 & -\\
				St.1 Se.6 &0.11 &0.35 & 0.11 &0.35 &0.50 &0.34 & -\\
				Te.1 Se.3 Te.3 & 0.23     & 0.20   & 0.28     & 0.19  & 0.56   & 0.20  & -\\
				\bottomrule
		\end{tabular}}
		\caption{Evaluations of \textit{SIFID}, \textit{SSIM} on \textit{Places50} \cite{Shaham2019SinGANLA}, \textit{LSUN50} \cite{Hinz2020ImprovedTF}, and \textit{ImageNet50}. St., Se. and Te. are the abbreviations of Structural, Semantic and Texture GAN. The numbers behind them are their stages.}
		\label{table1}
	\end{table*}

	\begin{table}[]
		\centering
		\setlength{\abovecaptionskip}{0.5em}
		\resizebox{\textwidth}{8mm}{
		{
			\begin{tabular}{@{}l|cc|cc|cc}
				\toprule
				\textbf{Model}  & \multicolumn{2}{c|}{\textit{\textbf{Places50}}} & \multicolumn{2}{c|}{\textit{\textbf{LSUN50}}} & \multicolumn{2}{c}{\textit{\textbf{ImageNet50}}}  \\
				\hline
				& Realism$\uparrow$    & Diversity$\uparrow$    & Realism$\uparrow$    & Diversity$\uparrow$  &  Realism$\uparrow$   & Diversity $\uparrow$ \\
				\hline
				SinGAN\cite{Shaham2019SinGANLA}    &23.0\%   & 47.8\% & 18.8\%   & 39.9\% & 11.5\%    & 35.3\%  \\

				ExSinGAN  & 77.0\% &  52.2\%  & 81.2\%    & 60.1\% &  88.5\% &  64.7\% \\
				\bottomrule
		\end{tabular}}}
		\caption{User studies on \textit{Places50}, \textit{LSUN50}, and \textit{ImageNet50}. Here the number represents the percentage of votes that voted for corresponding method. }
		\label{table2}
	\end{table}
	
	\noindent\textbf{Quantitative Evaluation.}
Constructing valid quantitative metric for single image generative model is really an open and difficult problem. We adopt the \textit{SIFID} proposed by SinGAN~\cite{Shaham2019SinGANLA}  to compare the distribution of the original image with that of syntheses using the features extracted by the pre-trained Inception network \cite{szegedy2015going}. The classic \textit{SSIM}~\cite{Wang2004ImageQA} is used for computing the diversity. Models  are evaluated on three datasets, \textit{Places50} proposed by SinGAN consisting of 50 images from the \textit{Places} dataset~\cite{zhou2014learning}, \textit{LSUN50} proposed by ConSinGAN consisting 50 images from the \textit{LSUN} dataset~\cite{yu2015lsun}, \textit{ImageNet50} containing 50 semantic images from the validation set of \textit{ImageNet} to test generation of semantic images. The quantitative evaluation in Table \ref{table1} shows that ExSinGAN achieves similar \textit{SSIM} to SinGAN but better performance on \textit{SIFID}. In particular, on the most difficult dataset \textit{ImageNet50}, ExSinGAN is ahead of other methods, proving that the hierarchical structure is highly effective to learn a pretty generative model. As previous discussion, DGP performs badly in texture (high \textit{SIFID}) but generates good layouts (low \textit{SSIM}). ExSinGAN takes both the strengths of GAN inversion (better layout) and internal learning methods (better pixel-level generation), performing outstandingly and stably all the time. \\

	\noindent\textbf{User Studies.}
Moreover, we conduct user studies to compare the realism and diversity of our syntheses with that of SinGAN via Amazon Mechanical Turk. We provide the original image and syntheses from ExSinGAN and SinGAN of the three datasets to Turkers, and Turkers will select the better image with regard to two criteria: realism (the more realistic image) and diversity (the image more different from the given image). Specifically, we recruited 50 Turkers for each dataset, and for each image we provide one synthesis of SinGAN and ExSinGAN respectively. The Turkers will make decisions without the time limitation. The result of user studies in Table \ref{table2} shows that our method can synthesize more realistic and diverse images on all datasets. Although the \textit{SSIM} of ExSinGAN is similar to that of SinGAN,  it can be attributed to that the syntheses of SinGAN contain artifacts that are meaningless to human vision. Moreover, as the difficulty of datasets increases, the gap between ExSinGAN and SinGAN also broadens, indicating that ExSinGAN has a better generalization performance on various images.

	\subsection{Generalization Study}
\noindent\textbf{Editing.} This task aims to produce a seamless composite in which image regions have been copied and pasted in other locations. For ExSinGAN, we inject a subsampled version of the naive image into stage one as $ x_{str} $, then combine the synthesis with the original image at the edited regions. Fig. \ref{syntheses}(a) shows some examples of three methods.  For the stone image, ExSinGAN performs better on generating more realistic texture than the other two methods. For the tree image, ExSinGAN can keep the structures of naive images well, which means ExSinGAN is more controllable. ConSinGAN failed to merge the background in this task.
	
	\noindent\textbf{Harmonization.} Image harmonization aims to realistically blend a pasted object with a background image. For SinGAN and ExSinGAN, the harmonization process is identical to the editing process.  The results in Fig. \ref{syntheses}(b) show that our work still gets better results compared with SinGAN. ConSinGAN can get better results sometimes because it fine-tunes the models by the naive image.
	
	\noindent\textbf{Paint-to-image.}
	See Fig. \ref{syntheses}(c), this task aims to transfer a paint into a photo-realistic image. Because ConSinGAN does not support this task, we just compare ExSinGAN with SinGAN. For SinGAN, this is done by subsampling the paint and feeding it into one of the coarse scales. For ExSinGAN, we just feed the subsampled painting into stage one. From these images, we know that ExSinGAN is sensitive to the structure of paint (the road under trees, the grass near the mountains), and the texture and style are more realistic (the cows). For the pyramids, the synthesis of ExSinGAN is more three-dimensional and context-based (the clouds around pyramids). In conclusion, ExSinGAN can do this task better than SinGAN with fewer stages and time consumption.  
	
	\subsection{Ablation Study}
	\begin{figure}[t]
		\begin{center}

			\begin{minipage}{0.6\linewidth}
				\begin{minipage}{0.15\linewidth}	
					{\includegraphics[width=1\textwidth]{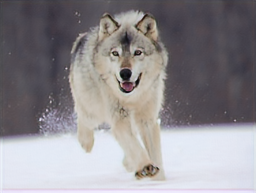}}
				\end{minipage}
				\begin{minipage}{0.15\linewidth}	
					{\includegraphics[width=1\textwidth]{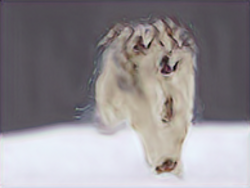}}
				\end{minipage}
				\begin{minipage}{0.15\linewidth}	
					{\includegraphics[width=1\textwidth]{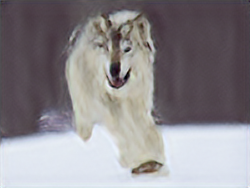}}
				\end{minipage}
				\begin{minipage}{0.15\linewidth}	
					{\includegraphics[width=1\textwidth]{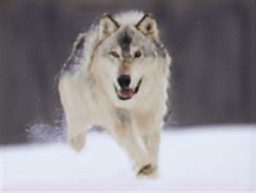}}
				\end{minipage}
				\begin{minipage}{0.15\linewidth}	
					{\includegraphics[width=1\textwidth]{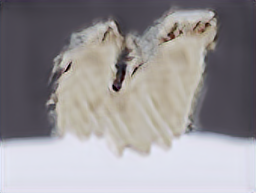}}
				\end{minipage}
				\begin{minipage}{0.15\linewidth}	
					{\includegraphics[width=1\textwidth]{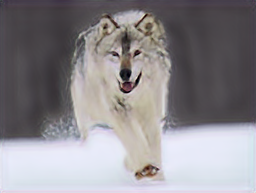}}
				\end{minipage}
				
				\begin{minipage}{0.15\linewidth}	
					\centerline{\includegraphics[width=1\textwidth]{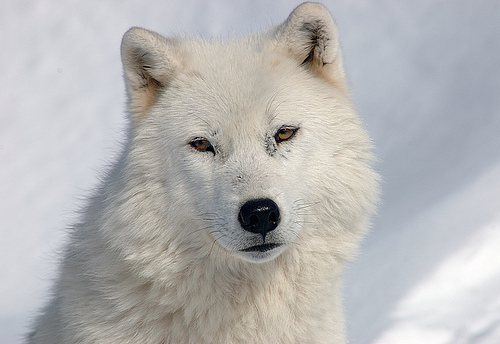}}
				\end{minipage}
				\begin{minipage}{0.15\linewidth}	
					\centerline{\includegraphics[width=1\textwidth]{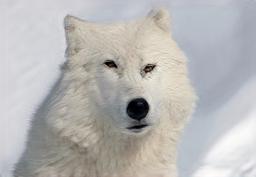}}
				\end{minipage}
				\begin{minipage}{0.15\linewidth}	
					\centerline{\includegraphics[width=1\textwidth]{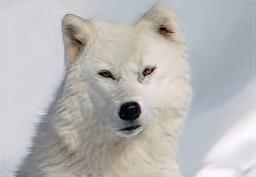}}
				\end{minipage}
				\begin{minipage}{0.15\linewidth}	
					\centerline{\includegraphics[width=1\textwidth]{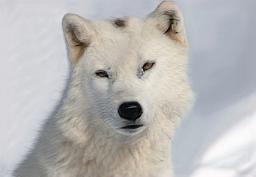}}
				\end{minipage}
				\begin{minipage}{0.15\linewidth}	
					\centerline{\includegraphics[width=1\textwidth]{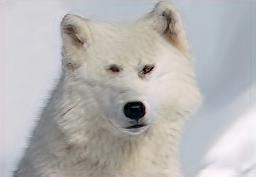}}
				\end{minipage}
				\begin{minipage}{0.15\linewidth}	
					\centerline{\includegraphics[width=1\textwidth]{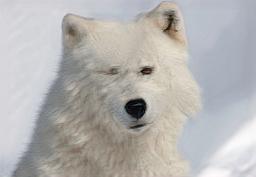}}
				\end{minipage}

				\begin{minipage}{0.15\linewidth}	
					\centerline{\includegraphics[width=1\textwidth]{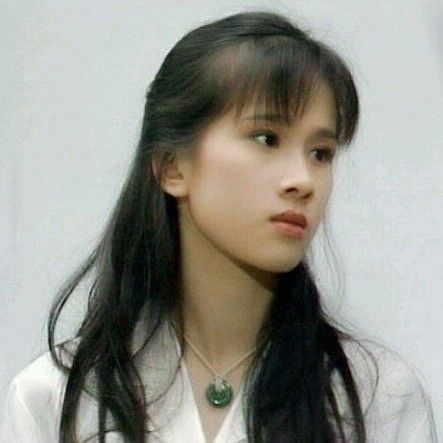}}
				\end{minipage}
				\begin{minipage}{0.15\linewidth}	
					\centerline{\includegraphics[width=1\textwidth]{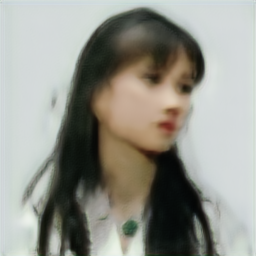}}
				\end{minipage}
				\begin{minipage}{0.15\linewidth}	
					\centerline{\includegraphics[width=1\textwidth]{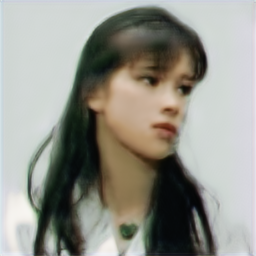}}
				\end{minipage}
				\begin{minipage}{0.15\linewidth}	
					\centerline{\includegraphics[width=1\textwidth]{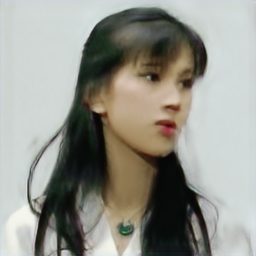}}
				\end{minipage}
				\begin{minipage}{0.15\linewidth}	
					\centerline{\includegraphics[width=1\textwidth]{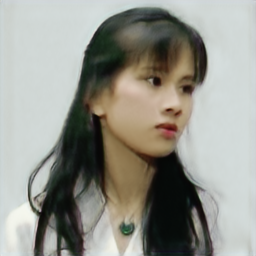}}
				\end{minipage}
				\begin{minipage}{0.15\linewidth}	
					\centerline{\includegraphics[width=1\textwidth]{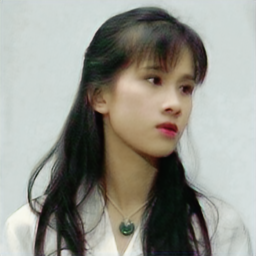}}
				\end{minipage}
				
			\end{minipage}
			
		\end{center}
		
		\caption{ Examples for ablation study. From left to right, first row: original, SinGAN, structural GAN with texture GAN, ExSinGAN, semantic GAN with texture GAN, using the discriminator of BigGAN for computing the perceptual loss. Second row: original, ExSinGAN with increasing $ n $ from 1 to 5. Third row: original, and ExSinGAN with increasing $ N $ from 5 to 9.}
		\label{places}
	\end{figure}
	
	In this subsection, we conduct both quantitative (Table \ref{table1}) and qualitative (Fig. \ref{places}) ablation study on ExSinGAN to verify that our improvements in SinGAN are effective. We start with the baseline 7-stage SinGAN  and check the effects of each modular GAN.
	
	\noindent\textbf{Structural GAN.} Fig. \ref{places} shows an awful synthesis of the baseline model. To prove the significance of structural GAN, we just replace the first stage of baseline with structural GAN, then the synthesis (the third image) has an overtly clearer appearance than before, and the \textit{SIFID} decreases overtly (St.1 Te.6 in Table \ref{table1}). 
	
	\noindent\textbf{Semantic GAN.} Now we combine the Structural GAN with Semantic GAN to show their effects further. In the first row of Fig. \ref{places}, the fourth image (ExSinGAN) shows that the skin of the wolf is smoother and tidier than the previous one. Then we withdraw the structural GAN (Te.1 Se.3 Te.3). The fifth image shows that the result of generation becomes terrible in vision. It means that the structural GAN is necessary for the successful generation, and the semantic GAN is a helper to make the synthesis more realistic and meaningful. They promote and affect each other.  In Table \ref{table1}, the quantitative results show that only Semantic GAN has a negative effect on the baseline, but a positive effect on structural GAN.  We also try to use the discriminator of BigGAN to compute the perceptual loss, and the results show that it plays a similar role with VGG-19, meaning it can extract useful semantic information.  
	
	\noindent\textbf{Number of Stages.} The number of stages greatly influences the results and the cost of time.  For ExSinGAN, there are two important hyper-parameters $ N $ and $ n $ that should be considered. When we fix $ N=6 $, the second row of Fig. \ref{places} shows that the semantics of syntheses is closer to the original image as $n$ is increased from 1 to 5. However, since the high-level feature maps also contain information for image restoration, once the size of the feature map is too large, the perceptual loss is almost equivalent to per-pixel loss,  which is inconsistent with the purpose of semantic constraint and adds artifacts to the syntheses. The quantitative results (St.1 Te.6, St.1 Se.1 Te.6, St.1 Se.3 Te.3, and St.1 Se.6) show that \textit{SIFID} firstly decreases and then increases, proving that $ n = 3 $ is the best choice. We do not conduct a quantitative study about  $ N $ because there has been similar research in~\cite{Shaham2019SinGANLA} and~\cite{Hinz2020ImprovedTF}. Here we just support some examples, the third row of  Fig. \ref{places} shows that higher-quality syntheses can be obtained with $ N $ increased, but of smaller changes of structure and more time consuming (a double-time when $ N = 9 $).  
	
	\section{Conclusion}
 	We introduced an explainable single image generative model named ExSinGAN based on the triplet representation of image. ExSinGAN has three modular GANs, respectively structure GAN generating reasonable layouts, semantic GAN replenishing the semantic details, and texture GAN adding the fine texture details. We demonstrated its superior abilities in synthesis and image manipulation tasks compared with previous works. We believe that the combination of external knowledge and internal learning is the key to solving the single image generation task, which will be further explored in our future work.

	\paragraph{Acknowledgements.} This paper is supported by the National Natural Science Foundation of China (Nos. U19B2040, 11731013, 11991022), the Strategic Priority Research Program of Chinese Academy of Sciences, Grant No. XDA27000000, and the Fundamental Research Funds for the Central Universities.
	\bibliography{egbib}
\end{document}